\documentclass[letterpaper,10pt,conference]{ieeeconf}

\IEEEoverridecommandlockouts

\usepackage{cite}
\usepackage{amsmath,amssymb,amsfonts}
\usepackage{graphicx}
\usepackage{textcomp}
\usepackage{xcolor}
\usepackage{cuted}
\usepackage{capt-of}
\usepackage{booktabs}
\usepackage{float}
\usepackage[hidelinks]{hyperref}
\hypersetup{
  pdftitle={ARSTAG: An Agentic Real2Sim2Real System for Task-Specific Robot Data Generation},
  pdfauthor={Bowei Li, Yuner Zhang, Changliu Liu}
}

\begin{document}

\title{\Large\bfseries
ARSTAG: An Agentic Real2Sim2Real System\\
for Task-Specific Robot Data Generation}

\author{%
{\normalsize Bowei Li, Yuner Zhang, and Changliu Liu}\\[0.2em]
{\small Carnegie Mellon University, Pittsburgh, PA 15213, USA}\\
{\small Email: \texttt{\{boweili, yunerz, cliu6\}@andrew.cmu.edu}}%
}

\maketitle

\begin{strip}
\centering
\includegraphics[width=1\textwidth]{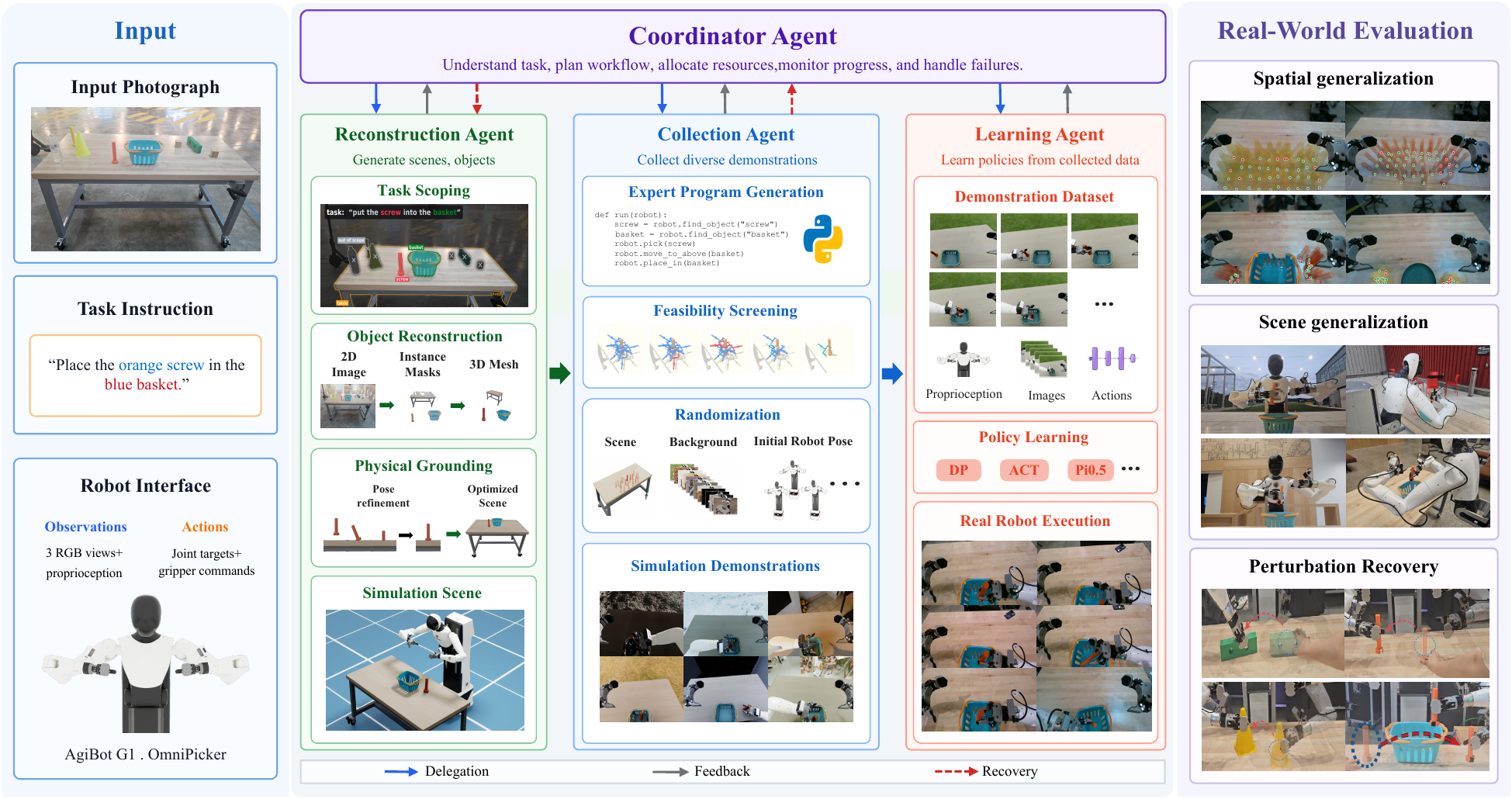}
\captionof{figure}{
    ARSTAG overview. Given a single RGB image and a task instruction, a coordinator guides three specialized agents through task-scoped reconstruction, demonstration generation, and policy learning for real-robot deployment, using tool feedback to coordinate revisions across stages. 
}
\label{fig:overview}
\end{strip}

\begin{abstract}
Adapting
visuomotor policies to new manipulation tasks often requires substantial
manual engineering or teleoperated data collection. Simulation can provide task-specific data at scale, but constructing the scene, designing expert behavior, and configuring data generation still require significant per-task effort. We present \textbf{ARSTAG}, an agentic Real2Sim2Real
system that turns a single RGB image and a natural-language instruction
directly into robot policy-learning data. A hierarchy of language agents
constructs a task-scoped simulation scene, generates robot-feasible demonstrations, and expands the
training distribution through task-consistent randomization, while a coordinator agent manages cross-stage feedback and recovery. Across seven manipulation tasks spanning grasping, placement,
and stacking, the ARSTAG-generated demonstrations enable sim-to-real
transfer of three visuomotor policy architectures to a dual-arm
robot, with $\pi_{0.5}$ achieving an average real-world success rate
of 74.6\%. Ablations show that task-consistent randomization substantially improves robustness, and policy performance increases with generated dataset size.
Project webpage: \url{https://boweili666.github.io/ARSTAG/}.

\end{abstract}

\section{Introduction}

In high-mix, low-volume settings, new products or task variants often require reconfiguring robot manipulation policies. Conventional integration relies on engineers to program robot manipulation behavior, and validate each task. Learning-based approaches reduce manual programming but still require substantial human effort for scene setup, teleoperated data collection, and repeated resets, especially when policies must generalize beyond a nominal configuration.

We study how to turn a new manipulation task into a deployable visuomotor policy with minimal human involvement. In a well-defined tabletop
manipulation setting, starting from only a
single RGB image of the workspace and a natural-language instruction, our goal is to automatically generate the task-specific data required for policy learning. The policy may be trained from scratch or obtained by post-training a pretrained generalist model. Our main focus is to automate the task-specific data-generation pipeline, while a dedicated learning agent manages downstream training and evaluation.

Two observations shape our approach. First,
task adaptation requires data, and simulation provides a scalable way to generate it without human demonstrations, since privileged access to ground-truth state in simulation sidesteps the
perception problems and could enable autonomous expert demonstration. In addition, prior sim-to-real and domain-randomization methods
show that randomized simulation supports efficient real-world transfer~\cite{tobin2017domain,peng2018sim}, while
recent real-to-sim-to-real systems and simulation data engines 
provide tools for scene reconstruction, demonstration generation, and
scalable evaluation~\cite{torne2024reconciling,han2025re,tao2024maniskill3}.

Second, applying these tools to a new task still requires
task-dependent decisions: 
what to reconstruct, which behavior to
generate, how to randomize the scene, and how to respond to failures. 
Language agents
offer a natural mechanism for coordinating these decisions~\cite{li2026roboclaw,xiao2026enpire}.
Advances in foundation models' multimodal reasoning, code generation,
and tool use create opportunities to automate task-specific robot
data generation. We use these capabilities for task and visual
grounding, expert-program generation, dataset preparation, and
policy-training configuration. We study how to combine them with
simulation, robot-feasibility checks, and task-consistent randomization
to generate training data, and evaluate the reconstructed scenes
and the real-world performance of policies trained on that data.

We introduce \textbf{ARSTAG}, an agentic Real2Sim2Real system that
generates task-specific policy-learning data from a single RGB image
and a natural-language instruction. 
As illustrated in Fig.~\ref{fig:overview}, a coordinator
agent orchestrates specialized reconstruction, collection, and learning agents, using tool feedback to coordinate revisions across stages. 
Extensive evaluation across ACT~\cite{zhao2023learning},
    Diffusion Policy~\cite{chi2025diffusion}, and
    $\pi_{0.5}$~\cite{intelligence2025pi_}
    showcases sim-to-real transfer
    on a dual-arm robot and robustness to the tested perturbations.
    Experiments demonstrate resilience to injected failures and
    reliable agent coordination, while ablations validate the effects
    of randomization, data scale, and reconstruction quality.

The main contributions of this work are:
\begin{itemize}
    \item \textbf{ARSTAG}, an agentic Real2Sim2Real system that
    generates policy-learning data from a single RGB image and
    instruction. A hierarchy of agents makes task-dependent tool-use decisions and uses tool results and error reports to coordinate revisions
within and across stages.

    \item A task-scoped reconstruction pipeline that grounds
    instructions in a scene graph, binds task objects to masks and
    3D assets, and geometrically repairs object poses to improve
    support consistency.

    \item A demonstration generator that grounds instructions in
    expert programs and combines feasibility screening, task-consistent
    randomization, and recovery demonstrations generated by a
    state-based expert after execution perturbations.

\end{itemize}

\section{Related Work}

\subsection{Post-Training Task Adaptation}
Post-training methods adapt pretrained policies to deployment tasks through
teleoperated demonstrations~\cite{kim2025fine},
human corrections~\cite{hu2026rac}, or real-world
rollouts~\cite{luo2025precise}. These approaches require human effort or robot
time in the deployment scene. ARSTAG instead generates task-specific data
autonomously in simulation from a single image and instruction, supporting
both post-training and training from scratch.

\subsection{Real-to-Sim Scene Construction and Data Generation}
Real-to-sim methods support object-level 3D asset
generation~\cite{chen2026sam}, scene
synthesis~\cite{yang2026sceneweaver,pfaff2026scenesmith},
and reusable environments reconstructed from real
observations~\cite{ranawaka2026simfoundry,zhang2026robosnap}.
Real-to-sim-to-real systems use these environments for robot-data
synthesis and transfer~\cite{torne2024reconciling,han2025re},
while simulation data engines provide scalable tasks,
demonstrations, randomization, and evaluation
protocols~\cite{mu2025robotwin,tao2024maniskill3}.
ARSTAG integrates task-scoped reconstruction into an end-to-end
data-generation pipeline, deriving both simulation scenes and
expert behavior from a single real image and a natural-language
instruction rather than requiring them to be prepared in advance.

\subsection{Agentic Orchestration for Robot Learning}

Language models have been used to map instructions to robot skills,
executable code, 3D affordances, and manipulation
constraints~\cite{ahn2022can,liang2023code},
while recent agentic systems organize long-horizon execution, recovery,
data collection, and policy improvement~\cite{li2026roboclaw,xiao2026enpire}.
These systems primarily use agents to execute or improve robot behavior.
ARSTAG instead uses agents to orchestrate an offline Real2Sim2Real
data-generation pipeline, where their decisions are evaluated through the
scenes, demonstrations, and training data they produce.

\section{Method}

We consider task-specific visuomotor policy learning for
tabletop manipulation. Given a single RGB image $I$ of the
workspace and a natural-language instruction $\ell$, ARSTAG
reconstructs a task-scoped simulation scene $\mathcal{S}$, generates
a demonstration dataset $\mathcal{D}$ at scale under task-consistent
randomization, and learns a policy $\pi$ from these demonstrations.
These three pipeline stages are handled by the reconstruction
agent, collection agent, and learning agent, respectively, under
the coordinator.
Reconstruction errors can compromise demonstration feasibility
and downstream policy learning. The coordinator therefore delegates
tasks, coordinates the handoff of intermediate outputs, and assigns
recovery tasks to the appropriate subagent.
Sections~\ref{sec:task_scoped_reconstruction}--\ref{sec:policy_learning}
describe each stage and its responsible subagent, while
Sec.~\ref{sec:hierarchical_agents} details their coordination
and recovery mechanisms.

\subsection{Task-Scoped Real-to-Sim Reconstruction}
\label{sec:task_scoped_reconstruction}
Given a single RGB image $I$ and task instruction $\ell$, the
reconstruction agent builds a task-scoped simulation scene
$\mathcal{S}$ through task grounding, visual grounding, and physical
grounding, followed by scene assembly and hand-off.

\textbf{Task grounding.}
To avoid reconstructing geometry unnecessary for the task, the agent
jointly reasons over $I$ and $\ell$ to construct a typed scene graph
$G=(V,E)$. It uses the instruction to select objects and supporting
surfaces, and the image to ground their visual attributes and spatial
relations. For the same image, ``pick up the screw'' selects the screw
and its supporting table, whereas ``place the screw in the basket''
also includes the basket. Each node $o_i\in V$ stores a class $c_i$ (e.g., screw, basket, or table)
and a short visual caption $d_i$ (e.g., ``yellow plastic screw'') for segmentation; each edge
$(o_i,r,o_j)\in E$ specifies a relation $r\in\mathcal{R}$ from a
predefined vocabulary of spatial and support relations. %

\textbf{Visual grounding.}
A caption may match multiple instances. Assigning the same mask
to multiple graph objects conflates distinct task-object identities,
potentially causing reconstruction and demonstration generation
to use the wrong instance. We therefore require partial one-to-one
matching between graph objects and masks. %
The agent prompts
SAM3~\cite{carion2026sam} with $d_i$ to obtain masks
$\{m_1,\dots,m_K\}$, then examines the input image $I$, the scene graph $G$, and a numbered mask
overlay to establish correspondences using visual attributes and
spatial relations. %
A validator enforces partial one-to-one
matching: each object and mask is used at most once.
For unmatched objects, the agent reviews the current candidates and
chooses whether to refine captions and repeat segmentation, invoke
segmentation with the generic prompt ``objects'', or revise assignments
over existing candidates. When another segmentation pass produces
additional masks, the agent combines them with the existing candidates
and repeats object--mask matching over the expanded set. %
Only assigned masks proceed to 3D reconstruction, and the agent
reports unresolved objects.

\textbf{Physical grounding.}
SAM~3D~\cite{chen2026sam} reconstructs each bound object from the image
and its mask, producing a mesh $M_i$, an estimated scale, and an
initial pose $T_i^{(0)}\in SE(3)$ in a camera-anchored frame.
Due to estimation errors, these estimates may leave a table tilted or an object floating above
its supporter. To address the problem, we introduce
a geometric pose repair method that prioritizes support consistency
over exact recovery of the observed layout.
Objects without a supporter in $G$, such as the table, are leveled
while preserving their heading and translated until their lower
bounds reach the ground plane.
Corrections are propagated to supported objects using
$T_i\leftarrow T_j(T_j^{(0)})^{-1}T_i^{(0)}$, where $i$ indexes the supported object, $j$ indexes its
supporter, $T_j$ is the supporter's current pose, and the
initial poses remain fixed. 
For a support pair whose current axis-aligned bounding boxes overlap
horizontally within tolerance $\delta$, vertical alignment uses
$
\Delta z_i = z^{\mathrm{top}}(B_j)+\delta-z^{\mathrm{bottom}}(B_i),
$
where $B_i$ and $B_j$ bound the supported object and its supporter,
$z$ denotes vertical height, and $\delta>0$ is a small clearance.
Propagation and alignment use bounded sweeps, stopping early when
no updates occur. Unlike scene refinement using rendered-view
feedback in SceneWeaver~\cite{yang2026sceneweaver} and optimization-based
collision repair in SceneSmith~\cite{pfaff2026scenesmith}, this repair
uses geometric transformations and bounding-box queries without
rendering or numerical pose optimization.

\textbf{Scene assembly and hand-off.}
The agent invokes scene assembly to turn reconstructed geometry into
an interactive simulation. An asset manifest links object identities to meshes and corrected
transforms. Assembly adds rigid-body physics and collision geometry,
then lets objects settle under gravity. Settling stops when all
objects' linear and angular speeds remain below rest thresholds
or when a step budget is reached to limit preparation time;
the resulting poses are then saved. %
Robot placement balances proximity to the table for manipulation
with clearance from the table and surrounding obstacles, while
favoring the camera-facing side. %
The scene is then expressed in the robot's base frame. 
The agent uses mask and asset checks to identify missing or near-empty
masks, absent meshes, degenerate transforms, and missing simulation-asset
references. It uses these diagnostics to guide bounded retries and
hands off the assembled scene with a summary of reconstructed objects
and unresolved mismatches.

\subsection{Task-Consistent Demonstration Generation}
\label{sec:demonstration_generation}

Given the reconstructed scene $\mathcal{S}$ and instruction $\ell$,
the collection agent prepares a state-based expert, screens its
grasp and motion targets, and expands the scene into a distribution
of task executions. Collection includes both nominal executions
and recovery from injected perturbations, retaining demonstrations
that satisfy task-specific success criteria.

\textbf{Task-conditioned expert programs.}
To turn the scene into robot behavior, the agent grounds $\ell$ in
Cartesian end-effector motions and gripper commands. Following ManiSkillFormer~\cite{yu2026maniskillformer} and
NeSyPack~\cite{li2025nesypack}, we express task behavior
as a program over object-state queries, end-effector
motion, and gripper control. For example, placing a screw in a basket comprises approach,
grasp, lift, transfer, and release, with motion targets computed
from the current screw and basket states. Access to simulator states
allows the same program to be instantiated across sampled layouts
without solving visual perception during expert execution.

\textbf{Grasp proposal generation.}
Grasp poses must account for both the reconstructed object geometry
and the robot's gripper. We therefore use
GraspGenX~\cite{han2026graspgen}, whose pretrained model conditions
on object point clouds and gripper geometry, reducing the need to
specify grasp poses manually for each object. The agent obtains
candidate gripper poses in the reconstructed reference scene and
may restrict them to a functional region, such as a screw shaft.

\textbf{Task-consistent domain randomization.}
The agent configures randomization to broaden the training
distribution while preserving task relations. Grasping targets
are sampled uniformly in $\mathcal{W}$. For pick-and-place,
the workspace can be partitioned into left, central, and right
regions, $\mathcal{W}=\mathcal{W}_L\cup\mathcal{W}_C\cup\mathcal{W}_R$,
with pick targets in the lateral regions and destinations in the
nominally shared central region.
Collection also varies object poses and distractor presence,
table size and offset, HDRI backgrounds and illumination,
camera intrinsics and extrinsics around nominal calibration,
and initial arm configurations and torso height.
To account for control-latency variation at deployment, each episode
samples $\tau\sim\mathcal{U}\{2,\dots,8\}$ and executes
$q_t^{\mathrm{exec}}=q_{t-\tau}^{\mathrm{cmd}}$ while recording
undelayed expert actions. The first target is held until the
delay buffer fills.

\textbf{Feasibility screening and layout resampling.}
Precomputed grasp candidates are transformed according to
the sampled pose of the target object, giving $\mathcal{H}=\{h_1,\dots,h_K\}\subset SE(3)$.
Among the tested subset $\mathcal{H}_{\mathrm{test}}\subseteq\mathcal{H}$,
accepted grasps satisfy
\begin{equation}
\mathcal{H}^{\star}=\{h\in\mathcal{H}_{\mathrm{test}}:
c_{\mathrm{scene}}(h)\wedge c_{\mathrm{geom}}(h)
\wedge c_{\mathrm{IK}}(h)\},
\label{eq:feasible_grasps}
\end{equation}
where $c_{\mathrm{scene}}$ checks gripper collisions against the
reference-scene point cloud, $c_{\mathrm{geom}}$ checks workspace,
body-envelope, and approach-clearance constraints, and
$c_{\mathrm{IK}}$ tests pre-grasp and grasp poses against IK error
tolerances. Program waypoints computable from the initial layout
also undergo position-based IK screening for the executing arm.

A failed grasp candidate does not by itself reject a layout.
If no tested grasp passes or a required program waypoint
fails its position-based IK check, 
the layout is rejected for collection, and the relevant object
poses are resampled within bounded attempts.
Persistent collection failures are reported to the coordinator,
which can request reconstruction corrections
(Sec.~\ref{sec:task_scoped_reconstruction}).

\textbf{Recovery demonstration generation.}
Visual randomization~\cite{tobin2017domain} and dynamics
randomization~\cite{peng2018sim} vary appearance and physical
response. We additionally perturb intermediate task-execution
states and use the state-based expert to generate corrective
continuations, explicitly pairing execution deviations with recovery
supervision. Two perturbations instantiate this design: a \emph{trajectory perturbation}
displaces the arm during approach, while a \emph{target perturbation}
induces a missed grasp by offsetting and tilting the planned grasp
pose before closing the gripper. Contact may displace the object,
which can also be relocated before recovery. In both cases, the
expert replans from the updated robot and object states to complete
the same task. 
The injected perturbation is excluded from supervision; only the
subsequent expert continuation is recorded and retained if it
satisfies the task-specific success criterion. This supplies
imitation learning with off-nominal observations paired with the
actions that resolve the deviation, without additional human
recovery demonstrations.
Appendix~\ref{app:expert-recovery} illustrates recovery data collection (Fig.~\ref{fig:supp-expert-recovery}).

Our focus is on generating a reusable task-specific dataset in a
single collection phase before policy training, with recovery
supervision obtained from perturbed expert executions. Future work
could further refine the learned policies through iterative data
aggregation with DAgger~\cite{ross2011reduction} or joint training
on simulated and real-world data.

\begin{figure*}[t]
    \centering
    \includegraphics[width=\textwidth]{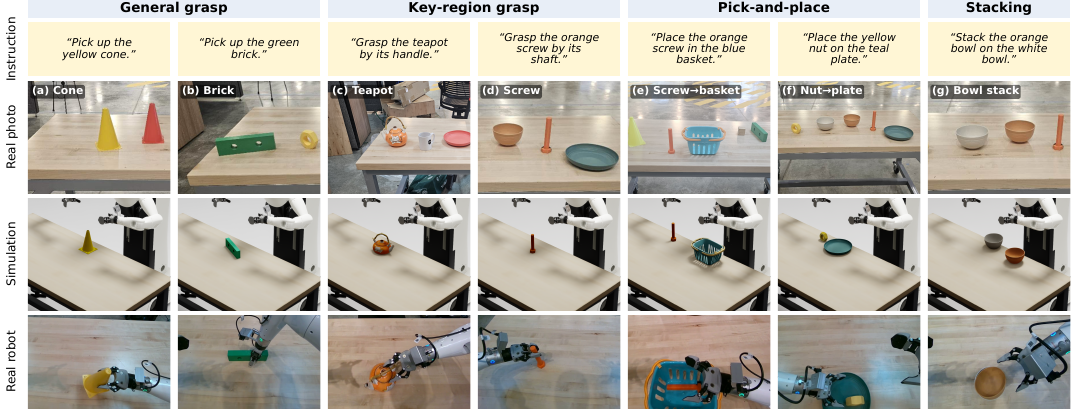}
    \caption{Overview of the seven tabletop manipulation tasks,
    grouped into general grasping, key-region grasping,
    pick-and-place, and stacking.
    From top to bottom, the rows show task instructions,
    real input photographs, task-scoped simulation scenes,
    and representative real-robot executions.}
    \vspace{-1em}
    \label{fig:tasks_overview}
\end{figure*}
\subsection{Policy Learning and Evaluation}
\label{sec:policy_learning}
Given demonstrations $\mathcal{D}$ and instruction $\ell$,
the learning agent configures policy-specific observation
and action interfaces and converts the data for training,
allowing different architectures to share the same source
demonstrations. It trains the policy $\pi$ and evaluates it
through closed-loop simulation rollouts under resampled
scene configurations, maintaining consistent interfaces
and preprocessing. Training and evaluation follow the
asynchronous mechanism in Sec.~\ref{sec:hierarchical_agents}.
The agent returns the checkpoint, interface settings,
and evaluation summary to the coordinator, together with
diagnostics if either job fails.

\subsection{Hierarchical Agent Coordination}
\label{sec:hierarchical_agents}
To support cross-stage revisions and long-running jobs,
we coordinate the three subagents through a coordinator,
role-specific system prompts, and an execution harness.

\textbf{Roles and execution constraints.}
We specify agent responsibilities, required inputs,
expected outputs, and recovery guidance in system prompts
and tool descriptions. Agents interpret task requirements
and tool feedback to determine the next operation and
select tool calls and arguments within this guidance.
The harness enforces role-specific tool access, validates
arguments, and limits tool-use rounds per turn.

\textbf{Feedback-driven coordination.}
The coordinator assigns tasks and provides each subagent
with the required inputs. 
The subagent reports the generated outputs and any failures,
and the coordinator uses this information to decide the
next step.
For example, our prompts prescribe object scale or
orientation adjustment and scene regeneration when
collection reports zero reachable grasps on a small
or thin object. Before collection is retried, the
coordinator delegates this correction to the reconstruction
agent, which uses the reported dimensions and grasp
diagnostics to select transformation parameters.
The prompts also instruct agents to stop retrying after
repeated identical failures and ask the user to choose
the target when the task requires a specific object
but the instruction and tool feedback leave multiple
candidates. Collection, training, and simulation evaluation run as
background jobs, allowing agents to end their turns after
launch; the execution framework then reports results or
failures to the coordinator, which decides how to proceed.

\section{Experiments}
We evaluate ARSTAG through four research questions:

\textbf{Q1 (Policy learning and transfer).}
Can demonstrations generated by ARSTAG support visuomotor
policy learning and sim-to-real transfer across
manipulation tasks?

\textbf{Q2 (Randomization and data scale).}
How do task-consistent randomization and demonstration
count (Sec.~\ref{sec:demonstration_generation}) affect
policy performance and robustness?

\textbf{Q3 (Reconstruction usability).}
Do identity binding and support repair
(Sec.~\ref{sec:task_scoped_reconstruction}) improve
the usability of reconstructed scenes?

\textbf{Q4 (Agent coordination).}
How do operating rules
(Sec.~\ref{sec:hierarchical_agents}) affect agent
coordination under injected tool failures?

For policy transfer and ablation studies
(Secs.~\ref{sec:policy_transfer} and~\ref{sec:ablation}),
task success rate is the primary evaluation metric.

\subsection{Experimental Setup}

\textbf{Platform and implementation.}
We use an AgiBot G1 robot with two 7-DoF arms,
each equipped with an OmniPicker gripper. Observations are provided
by one head-mounted camera and one wrist-mounted camera per arm.
Demonstrations are recorded and policies are executed at 10\,Hz.
Scene assembly and demonstration generation use NVIDIA Isaac
Sim, with camera viewpoints and observation
modalities corresponding to the real platform. GPT-5.5 is used
for agent decision-making and multimodal reasoning.
SAM3~\cite{carion2026sam} provides open-vocabulary segmentation,
SAM3D~\cite{chen2026sam} reconstructs individual objects, and
GraspGenX~\cite{han2026graspgen} generates grasp proposals. Demonstration collection runs on an NVIDIA GeForce RTX 5080
Laptop GPU, while policy training uses NVIDIA L20 GPUs.

\textbf{Policy training.}
For the main sim-to-real experiments, all policies use the same source demonstrations for each task.
ACT~\cite{zhao2023learning} and DP~\cite{chi2025diffusion}
are trained for 200\,000 steps with batch size 128.
To account for differences in robot embodiment and experimental
setup, we adapt $\pi_{0.5}$~\cite{intelligence2025pi_} to our
platform by LoRA-fine-tuning \texttt{pi05\_base} on
ARSTAG-generated simulation demonstrations for 30\,000 steps
with batch size 32. All reported $\pi_{0.5}$ results use
the fine-tuned policy. We deploy the final checkpoints without early stopping
or validation-based checkpoint selection.
Full training configurations are provided in Appendix~\ref{app:training} (Table~\ref{tab:supp-training}).

\textbf{Tasks and demonstrations.}
We consider seven tabletop manipulation tasks covering general
grasping, key-region grasping, pick-and-place, and stacking
(Fig.~\ref{fig:tasks_overview}). Each task uses a
real RGB image and a natural-language instruction.
For each task in the real-robot transfer study, we collect
1000 successful demonstrations in simulation using the
procedure in Sec.~\ref{sec:demonstration_generation}.
Mean demonstration durations range from 11.3 to 11.9\,s
for grasping tasks and from 17.3 to 19.6\,s for
pick-and-place-style tasks, including bowl stacking.

\textbf{Real-robot evaluation.}
Each reported task--policy pair is evaluated over 40 real-robot
trials. Object placements are resampled between trials following
the task-specific sampling rules used for collection. Grasping targets are
sampled uniformly over the workspace; for tasks using regional
sampling, each object is sampled within its assigned region. The table is placed in front of the
robot without precise calibration of its relative pose, and its
position and orientation vary across trials.
For nut-to-plate placement, all real-robot trials start with the nut
upright, since the state-based expert finds no suitable grasp pose for
the flat nut in the reconstructed scene. Flat initial poses are not
evaluated. Non-prehensile manipulation to reorient objects before
grasping is outside the scope of this work.
A trial succeeds
when its final state satisfies the task-specific criterion,
such as object lifting, placement, or stacking.
Appendix~\ref{app:evaluation-sampling} visualizes the sampled initial placements and all 40 cone-grasping trials (Figs.~\ref{fig:supp-trial-grid}--\ref{fig:supp-placement-map}).

\subsection{Policy Learning and Sim-to-Real Transfer}
\label{sec:policy_transfer}
\begin{table}[t]
\caption{Real-robot success rates (\%) on seven tasks.}
\label{tab:main_results}
\centering
\small
\setlength{\tabcolsep}{7pt}
\renewcommand{\arraystretch}{1.1}
\begin{tabular}{l|ccc}
\hline
Task & ACT & DP & $\pi_{0.5}$ \\
\hline
Cone             & 55   & \textbf{77.5} & 65 \\
Brick            & 25   & 57.5          & \textbf{92.5} \\
Teapot           & 40   & 67.5          & \textbf{72.5} \\
Screw            & 62.5 & 72.5          & \textbf{90} \\
Screw$\to$basket  & 42.5 & \textbf{62.5} & 60 \\
Nut$\to$plate    & 35   & 62.5          & \textbf{67.5} \\
Bowl stacking       & 45   & 67.5          & \textbf{75} \\
\hline
Average          & 43.6 & 66.8          & \textbf{74.6} \\
\hline
\end{tabular}
\end{table}

\begin{figure}[t]
    \centering
    \includegraphics[width=\columnwidth]
    {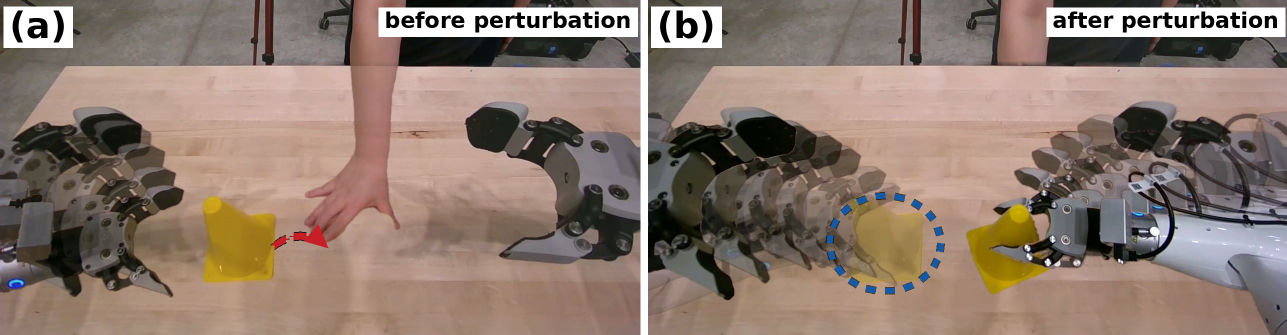}
    \caption{Real-robot recovery during cone grasping:
(a) before and (b) after manual target displacement.
The red arrow indicates the displacement direction,
and the blue dashed circle marks the original target position.
Transparent overlays show successive gripper poses.}
    \vspace{-1em}
    \label{fig:trial_map}
    \label{fig:real_recovery}
\end{figure}

\textbf{Sim-to-real transfer.}
The generated demonstrations support real-robot policy
execution across seven tasks spanning general grasping,
key-region grasping, pick-and-place, and stacking
(Fig.~\ref{fig:tasks_overview}; Table~\ref{tab:main_results}).
Diffusion Policy achieves 57.5--77.5\% success,
exceeding ACT by 10--32.5 percentage points across all
seven tasks. Fine-tuned $\pi_{0.5}$ achieves the highest
observed success on brick grasping (92.5\%), teapot-handle
grasping, screw-shaft grasping (90\%), nut placement,
and bowl stacking (75\%), while Diffusion Policy performs
best on cone grasping and screw placement in a basket.
These results show that the generated datasets support
both task-specific policy training and pretrained
vision-language-action policy adaptation for real-robot
execution.

\textbf{Recovery on the real robot.}
Fig.~\ref{fig:real_recovery} illustrates recovery
from target displacement using a policy trained with
the full demonstration-generation recipe.
During approach, an operator moves the cone to another
position before gripper contact. The policy continues
without a reset, retracts the initially reaching arm,
and completes the grasp with the other arm at the
updated target position.
This example illustrates closed-loop adaptation to
a change in object position during real execution
and complements the quantitative mis-grasp recovery
evaluation in simulation (Sec.~\ref{sec:ablation}).
Additional real-world recovery sequences are shown in Appendix~\ref{app:real-recovery} (Fig.~\ref{fig:supp-real-recovery}).

\textbf{Failure cases and limitations.}
Small-block stacking and screw insertion remain challenging
in exploratory trials, potentially due to high alignment
requirements, sim-to-real discrepancies, and asymmetric
gripper geometry. The generated data also does not reliably
support learning transitions between successive object
manipulations in longer pick-and-place sequences.
Future work will address precision manipulation and
long-horizon task composition.

\subsection{Ablation of Randomization and Data Scale}
\label{sec:ablation}

\begin{table}[t]
\caption{Randomization ablation: success rates (\%).}
\label{tab:randomization}
\centering
\small
\setlength{\tabcolsep}{5pt}
\begin{tabular}{llcc}
\toprule
Randomization & Condition & Full & w/o \\
\midrule
Execution  & none         & \textbf{82.5} & \textbf{82.5} \\
           & in-dist.     & \textbf{77.5} & 72.5 \\
           & out-of-dist. & \textbf{75.0} & 40.0 \\
\midrule
Geometry   & none         & 72.5 & \textbf{87.5} \\
           & in-dist.     & \textbf{77.5} & 70.0 \\
           & out-of-dist. & \textbf{62.5} & 40.0 \\
\midrule
Background & none         & 72.5          & \textbf{87.5} \\
           & in-dist.     & \textbf{77.5} & 0.0 \\
           & out-of-dist. & \textbf{70.0} & 0.0 \\
\bottomrule
\vspace{-2em}
\end{tabular}
\end{table}
\textbf{Randomization ablation.}
We perform leave-one-out ablations on \emph{grasp the screw}
in simulation, holding reconstruction, the collector,
Diffusion Policy architecture, and training budget fixed.
Each dataset is recollected without one randomization
family: execution (control delay and trajectory/target
perturbations), geometry (table size/offset and torso
height), or background (HDRI).
Evaluation varies the removed family while keeping the
others enabled, with 40 held-out layouts per cell
(Table~\ref{tab:randomization}).
Execution and geometry are tested without perturbation,
within the collection range (ID), and beyond it (OOD).
Execution OOD induces a failed grasp followed by object
displacement to test re-grasping; geometry OOD varies
table scale, offset, yaw, and torso height beyond their
collection ranges.
Background evaluation uses the static scene background,
the 539-map collection pool (ID), and a disjoint pool
of 456 predominantly outdoor HDRIs (OOD).

Without perturbing the tested family, the execution variants both
achieve 82.5\%, whereas removing geometry or background randomization
increases success from 72.5\% to 87.5\% in the corresponding
condition. Under ID perturbations, the full recipe achieves 77.5\%
in all three comparisons, versus 72.5\%, 70.0\%, and 0.0\% without
execution, geometry, and background randomization, respectively.
Under OOD conditions, the full recipe achieves 75.0\% versus 40.0\%
in the mis-grasp recovery test, and 62.5\% versus 40.0\% under geometry
variation. The geometry and background comparisons reveal a trade-off
between performance under fixed conditions and robustness to the
corresponding variations. %

\textbf{Data scale.}
We train Diffusion Policy on nested subsets of
50, 100, 200, 500, and 1000 demonstrations from the same
dataset for \emph{grasp the screw}, using 100\,000 training
steps and 40 held-out simulation episodes per policy.
Success rates are 0\%, 7.5\%, 27.5\%, 52.5\%, and 62.5\%,
respectively. Doubling the demonstration count from 500
to 1000 yields only a 10-percentage-point improvement,
indicating diminishing returns per additional demonstration.
We therefore adopt 1000 demonstrations per task as a
practical budget balancing performance and data volume.
Appendix~\ref{app:data-scale} plots these results (Fig.~\ref{fig:supp-data-scale}).  %

\begin{figure}[t]
    \centering
    \includegraphics[width=0.75\columnwidth]
    {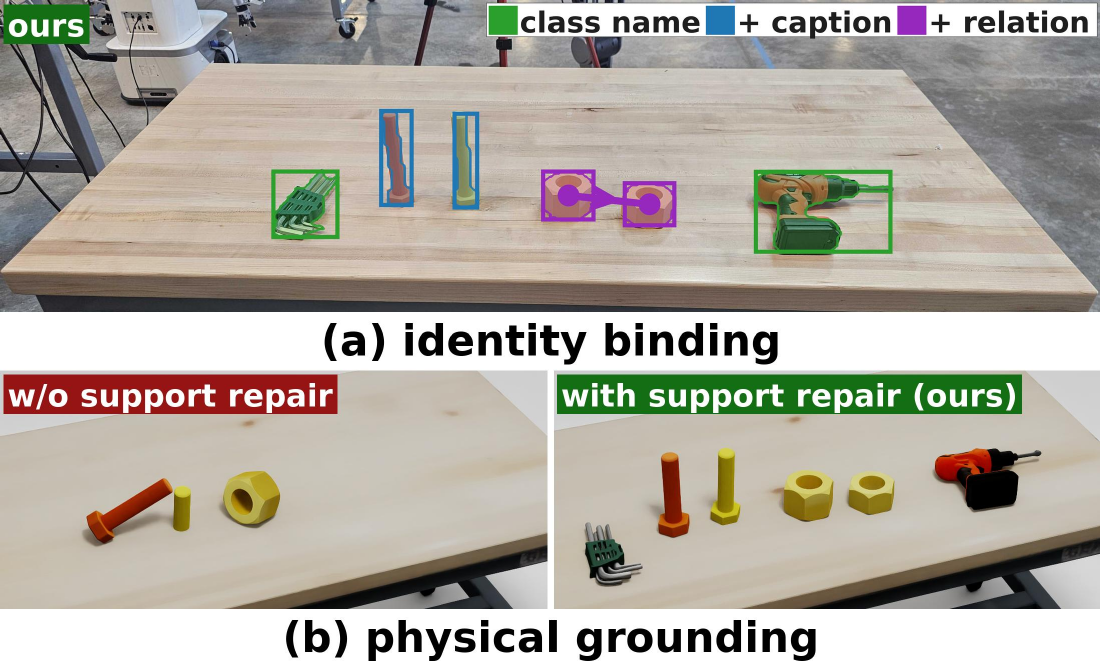}
    \caption{Identity binding and support repair.
    (a) Class names distinguish the hex keys and drill
    (green), captions distinguish the screws by color
    (blue), and relative size together with the
    ``left of'' relation distinguishes the two nuts
    (purple).
    (b) The same reconstructed scene under gravity
    without (left) and with (right) support repair.
    Without repair, three of the six objects lie below
    the table surface and are not visible.}
    \vspace{-2em}
    \label{fig:recon_usability}
\end{figure}

\subsection{Task-Scoped Reconstruction and Scene Usability}
\label{sec:recon_usability}

\textbf{Task-scoped versus full-tabletop reconstruction.}
We compare 42 task-scoped scenes with 12 full-tabletop
reconstructions from the same 12 photographs.
The task-scoped scenes contain 125 graph objects and
83 support relations; their instructions extend beyond
the seven real-robot tasks and include different object
selections from the same photograph.
Task scoping reduces the mean number of generated meshes
from 7.8 to 3.0 (62\% fewer) and the median number of
candidate masks from 12 to 3 (Table~\ref{tab:task_scoping}).
During the same overnight session on the same server,
four additional task-scoped runs on three photographs
take 4.9--7.6 minutes, compared with 21.7--30.8 minutes
for the corresponding full-tabletop runs, yielding
a 3.1--6.1$\times$ speedup per task.
Unmatched graph objects number 0/125 and 4/97 in the
task-scoped and full-tabletop scenes, respectively.
\begin{table}[t]
\caption{Task scoping and reconstruction usability.}
\label{tab:task_scoping}
\label{tab:recon}
\centering
\footnotesize
\setlength{\tabcolsep}{4pt}
\renewcommand{\arraystretch}{1.05}
\begin{tabular}{@{}lcc@{}}
\toprule
\textbf{Task scoping} & Full-tabletop & Task-scoped \\
\midrule
Generated meshes, mean
    & 7.8 & \textbf{3.0} \\
Candidate masks, median
    & 12 & \textbf{3} \\
Reconstruction time (min)$^\dagger$
    & 21.7--30.8 & \textbf{4.9--7.6} \\
Unmatched graph objects
    & 4/97 & \textbf{0/125} \\
\midrule
\textbf{Identity binding} & Class name & Ours \\
\midrule
Unique in the scene (61)
    & 100\% & \textbf{100\%} \\
Class matches a non-task object (25)
    & 8\% & \textbf{100\%} \\
Repeated class, captions differ (22)
    & 36\% & \textbf{100\%} \\
Repeated class, same color (17)
    & 29\% & \textbf{100\%} \\
All objects (125)
    & 61\% & \textbf{100\%} \\
\midrule
\textbf{Physical grounding} & w/o repair & Ours \\
\midrule
Support gap, median (mm)
    & 34.4 & \textbf{0.1} \\
Contact rate (83 relations)
    & 2\% & \textbf{100\%} \\
Displacement, median (mm)
    & 23.3 & \textbf{8.8} \\
Displacement, p90 (mm)
    & 646.6 & \textbf{24.9} \\
\bottomrule
\end{tabular}

\smallskip
\parbox{\columnwidth}{\scriptsize
$^\dagger$Ranges over four task-scoped runs and three
full-tabletop runs from three photographs.
Each full-tabletop run is shared by tasks using that
photograph. Timing excludes scene-graph generation
and simulator scene assembly.
}
\vspace{-2em}
\end{table}

\textbf{Identity binding.}
Identity binding links task-selected objects and supporting
surfaces to their intended image instances using class,
appearance, and spatial cues
(Fig.~\ref{fig:recon_usability}(a)).
We compare our procedure with a class-name baseline
that selects the first matching mask.
Of the 125 objects, 64 have multiple class-matching
candidates. Ambiguous assignments are scored against
ground truth from numbered-mask overlays; single-candidate
objects are counted as correct for both methods.
The baseline achieves 61\% overall accuracy, with at least
one incorrect binding in 28/42 scenes, whereas our procedure
achieves 100\% in each group on this evaluation set
(Table~\ref{tab:recon}).
Incorrect bindings can omit required objects, select
unintended instances, or misidentify supporting surfaces,
compromising demonstration generation.
Additional identity-binding examples appear in Appendix~\ref{app:binding} (Fig.~\ref{fig:supp-binding}). 
\begin{figure}[t]
\centering
\includegraphics[width=1\linewidth]
{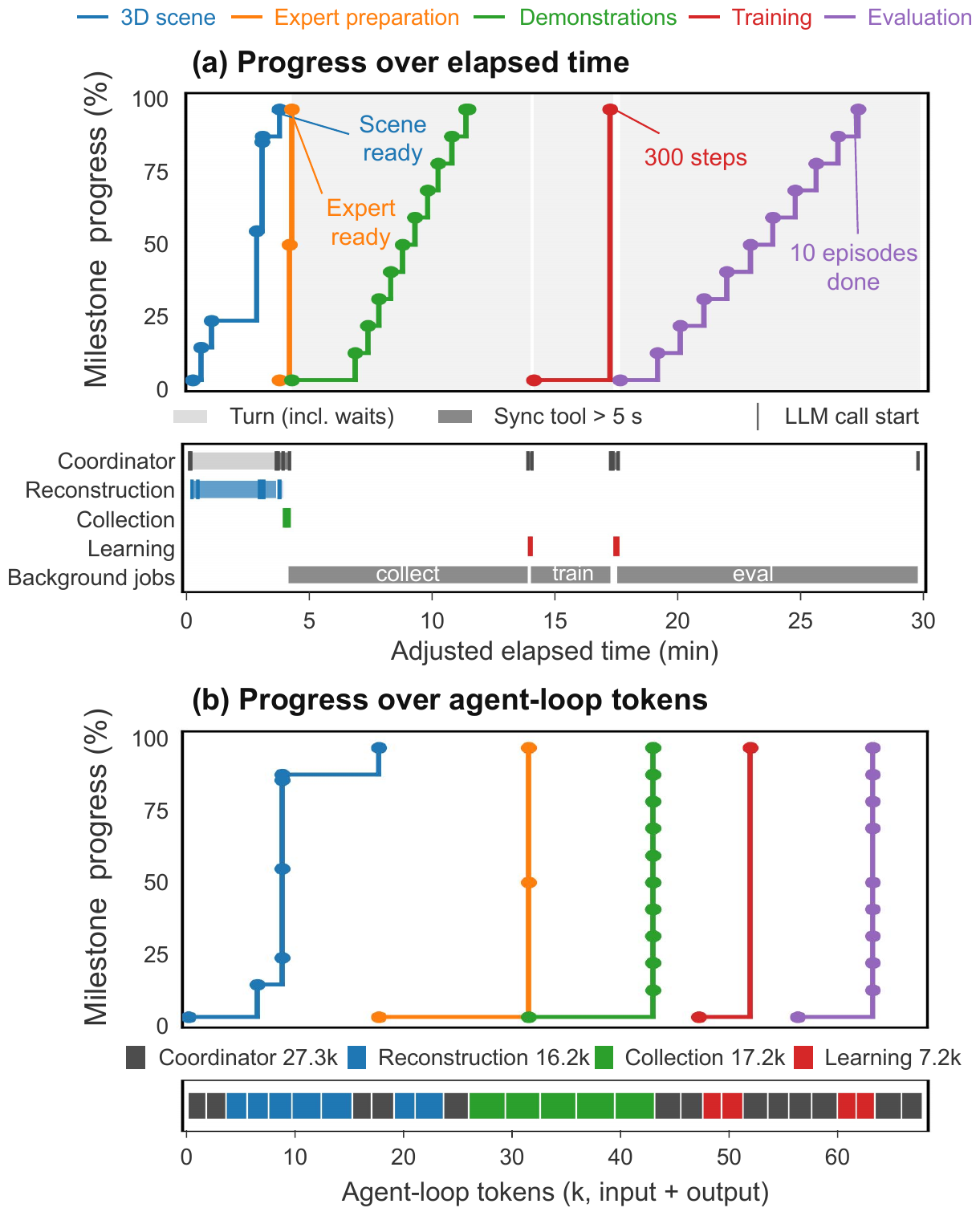}
\caption{Agent execution profile for screw grasping. (a) Pipeline progress and agent activity over elapsed time; gray regions denote asynchronous jobs. (b) Progress versus cumulative agent-loop tokens. Colors denote agents; bars indicate agent turns and synchronous tool calls. 
}
\vspace{-1em}
\label{fig:agent_profile}
\end{figure}

\begin{figure*}[t]
\centering
\includegraphics[width=1\textwidth]
{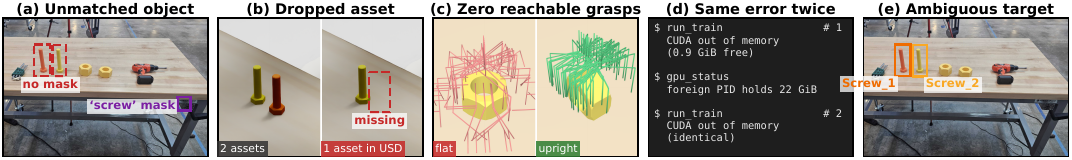}
\caption{Failure cases for agent coordination:
(a) unmatched target;
(b) omitted asset;
(c) no reachable grasps, with reorientation shown as a
possible correction (red: colliding; green: collision-free
candidates);
(d) repeated GPU out-of-memory errors; and
(e) ambiguous target.}
\vspace{-1.5em}
\label{fig:injected_failures}
\end{figure*}

\textbf{Physical grounding.}
The \emph{w/o repair} condition omits only the final
vertical support alignment, retaining root leveling,
ground alignment, and correction propagation
(Sec.~\ref{sec:task_scoped_reconstruction}).
Contact requires the absolute vertical gap between the
supported object's bounding-box bottom and its supporter's
bounding-box top to be within 1\,mm.
Repair reduces the median gap from 34.4 to 0.1\,mm
and increases the contact rate from 2\% to 100\%.
Starting from scenes before settling, we simulate
300 physics steps under gravity for the 83 supported
objects. Repair reduces median displacement from
23.3 to 8.8\,mm and the 90th percentile from
646.6 to 24.9\,mm
(Table~\ref{tab:recon}; Fig.~\ref{fig:recon_usability}(b)).
Residual motion remains possible, such as an object
sliding inward from a concave support's rim.
Appendix~\ref{app:support} provides further settling comparisons (Fig.~\ref{fig:supp-support}).

\subsection{Agent Coordination Under Injected Failures}
\label{sec:harness_eval}
\textbf{Agent execution profile.}
We profile a screw-grasping run with 10 successful
demonstrations, 300 training steps, and 10 closed-loop
simulation evaluation episodes.
The run collects the demonstrations in 10 attempts,
recording 29 agent-loop LLM calls, 67.8k input and
output tokens, and 22 tool calls.
Expert preparation includes key-region annotation
and expert program generation.
Collection, training, and evaluation run asynchronously:
stage agents return job identifiers and end their turns,
while completion events notify the coordinator to
initiate subsequent stages.
Pipeline progress, agent activity, and agent-loop token
usage are summarized in Fig.~\ref{fig:agent_profile}.
Token counts exclude model calls internal to tools.

\textbf{Controlled coordination stress test.}
We evaluate the implemented GPT-5.5 agent loop using
scripted text responses representing five observed tool
failures (Fig.~\ref{fig:injected_failures}).
We compare the full configuration with a variant without
prompt-level operating rules and a fixed stage chain that
retries each failed step once, using ten trials per failure
and configuration.
Trials are assessed from tool-call traces and final
responses against prescribed case-specific criteria:
bounded retry followed by reporting and proceeding,
assembly verification and repetition, revision by the
reconstruction agent, stopping and reporting a repeated
error, or requesting clarification.
The full configuration meets these criteria in 50/50
trials, compared with 23/50 without operating rules and
10/50 for the fixed chain.
These results support the role of operating rules in
guiding failure handling; meeting the criteria includes
appropriate stopping or clarification rather than
necessarily completing the pipeline.

\textbf{Qualitative recovery with real tool execution.}
We compare ARSTAG with a fixed pipeline using the same
photograph, instruction, and underlying tools for
\emph{placing the yellow nut on the teal plate}.
For the reconstructed flat nut, the configured grasp planner initially
finds no collision-free candidates, and the fixed pipeline produces no
successful demonstrations after retrying collection.
Following the diagnostics and prescribed recovery
guidance, the coordinator delegates scene revision to
the reconstruction agent.
The agent enlarges the simulated nut by $1.5\times$,
rotates it by $90^\circ$ to an upright pose, sets its mass to 0.02\,kg,
and corrects support penetration before rebuilding and
verifying the scene.
The revised scene yields 897 collision-free grasp
candidates and a successful placement demonstration
on the second rollout, illustrating recovery of demonstration
generation through cross-stage scene revision.
The recorded workflow is summarized in Appendix~\ref{app:agent-recovery} (Figs.~\ref{fig:supp-agent-recovery}--\ref{fig:supp-agent-visuals}).

\section{Conclusion}

We presented ARSTAG, an agentic Real2Sim2Real system that turns a
single RGB image and a natural-language instruction into robot
policy-learning data. A coordinator agent and three stage agents scope
reconstruction to the task, ground the instruction in robot-feasible
behavior, and expand a single observation into a randomized training
distribution, so that decisions previously requiring per-task
engineering are made by the agents. Policies trained on the generated demonstrations transfer
to seven real-robot manipulation tasks, with $\pi_{0.5}$
achieving an average success rate of 74.6\%.
Our ablations show that each randomization family chiefly
improves robustness to the perturbation family it models.

\section*{Acknowledgments}
ChatGPT assisted with manuscript writing and editing, and
Claude Code with implementation of the system described
in Secs.~\ref{sec:task_scoped_reconstruction}--\ref{sec:hierarchical_agents}.
The authors take full responsibility for the manuscript
and implementation.

\bibliographystyle{IEEEtran}
\bibliography{references}

@inproceedings{chen2026sam,
  title={Sam 3d: 3dfy anything in images},
  author={Chen, Xingyu and Chu, Fu-Jen and Gleize, Pierre and Liang, Kevin J and Sax, Alexander and Tang, Hao and Wang, Weiyao and Guo, Michelle and Hardin, Thibaut and Li, Xiang and others},
  booktitle={Proceedings of the IEEE/CVF Conference on Computer Vision \& Pattern Recognition},
  pages={7220--7232},
  year={2026}
}

@article{yang2026sceneweaver,
  title={Sceneweaver: All-in-one 3d scene synthesis with an extensible and self-reflective agent},
  author={Yang, Yandan and Jia, Baoxiong and Zhang, Shujie and Huang, Siyuan},
  journal={Advances in neural information processing systems},
  volume={38},
  pages={140319--140351},
  year={2026}
}

@inproceedings{pfaff2026scenesmith,
  title={Scenesmith: Agentic generation of simulation-ready indoor scenes},
  author={Pfaff, Nicholas and Cohn, Thomas and Zakharov, Sergey and Cory, Rick and Tedrake, Russ},
  booktitle={Forty-third International Conference on Machine Learning},
  year={2026}
}

@article{ranawaka2026simfoundry,
  title={Simfoundry: Modular and automated scene generation for policy learning and evaluation},
  author={Ranawaka, Nadun and Wong, Josiah and Pai, Wei-Lin and Chu, Wei-Teng and Dai, Tianyuan and Moghani, Masoud and Yin, Hang and Jiang, Yunfan and Durbano, Wesley and Huynh, Brandon and others},
  journal={arXiv preprint arXiv:2606.28276},
  year={2026}
}

@article{zhang2026robosnap,
  title={RoboSnap: One-Shot Real-to-Sim Scene Generation for Generalizable Robot Learning and Evaluation},
  author={Zhang, Shujie and Yi, Jingkun and Zhong, Weipeng and Zhou, Zirui and Zhu, Yangkun and Wang, Hanqing and Xu, Xudong and Zhang, Weinan and Shen, Chunhua},
  journal={arXiv preprint arXiv:2607.06699},
  year={2026}
}

@inproceedings{tobin2017domain,
  title={Domain randomization for transferring deep neural networks from simulation to the real world},
  author={Tobin, Josh and Fong, Rachel and Ray, Alex and Schneider, Jonas and Zaremba, Wojciech and Abbeel, Pieter},
  booktitle={2017 IEEE/RSJ international conference on intelligent robots and systems (IROS)},
  pages={23--30},
  year={2017},
  organization={IEEE}
}

@inproceedings{peng2018sim,
  title={Sim-to-real transfer of robotic control with dynamics randomization},
  author={Peng, Xue Bin and Andrychowicz, Marcin and Zaremba, Wojciech and Abbeel, Pieter},
  booktitle={2018 IEEE international conference on robotics and automation (ICRA)},
  pages={3803--3810},
  year={2018},
  organization={IEEE}
}

@article{torne2024reconciling,
  title={Reconciling reality through simulation: A real-to-sim-to-real approach for robust manipulation},
  author={Torne, Marcel and Simeonov, Anthony and Li, Zechu and Chan, April and Chen, Tao and Gupta, Abhishek and Agrawal, Pulkit},
  journal={arXiv preprint arXiv:2403.03949},
  year={2024}
}

@article{han2025re,
  title={Re\textsuperscript{3} Sim: Generating High-Fidelity Simulation Data via 3D-Photorealistic Real-to-Sim for Robotic Manipulation},
  author={Han, Xiaoshen and Yu, Junqiu and Liu, Minghuan and Chen, Yilun and Lyu, Xiaoyang and Tian, Yang and Wang, Bolun and Zhang, Weinan and Pang, Jiangmiao},
  journal={arXiv preprint arXiv:2502.08645},
  year={2025}
}

@inproceedings{mu2025robotwin,
  title={Robotwin: Dual-arm robot benchmark with generative digital twins},
  author={Mu, Yao and Chen, Tianxing and Chen, Zanxin and Peng, Shijia and Lan, Zhiqian and Gao, Zeyu and Liang, Zhixuan and Yu, Qiaojun and Zou, Yude and Xu, Mingkun and others},
  booktitle={2025 IEEE/CVF Conference on Computer Vision and Pattern Recognition (CVPR)},
  pages={27649--27660},
  year={2025},
  organization={IEEE}
}

@article{tao2024maniskill3,
  title={Maniskill3: Gpu parallelized robotics simulation and rendering for generalizable embodied ai},
  author={Tao, Stone and Xiang, Fanbo and Shukla, Arth and Qin, Yuzhe and Hinrichsen, Xander and Yuan, Xiaodi and Bao, Chen and Lin, Xinsong and Liu, Yulin and Chan, Tse-kai and others},
  journal={arXiv preprint arXiv:2410.00425},
  year={2024}
}

@article{ahn2022can,
  title={Do as i can, not as i say: Grounding language in robotic affordances},
  author={Ahn, Michael and Brohan, Anthony and Brown, Noah and Chebotar, Yevgen and Cortes, Omar and David, Byron and Finn, Chelsea and Fu, Chuyuan and Gopalakrishnan, Keerthana and Hausman, Karol and others},
  journal={arXiv preprint arXiv:2204.01691},
  year={2022}
}

@inproceedings{liang2023code,
  title={Code as policies: Language model programs for embodied control},
  author={Liang, Jacky and Huang, Wenlong and Xia, Fei and Xu, Peng and Hausman, Karol and Ichter, Brian and Florence, Pete and Zeng, Andy},
  booktitle={2023 IEEE International conference on robotics and automation (ICRA)},
  pages={9493--9500},
  year={2023},
  organization={IEEE}
}

@article{li2026roboclaw,
  title={Roboclaw: An agentic framework for scalable long-horizon robotic tasks},
  author={Li, Ruiying and Zhou, Yunlang and Zhu, YuYao and Chen, Kylin and Wang, Jingyuan and Wang, Sukai and Hu, Kongtao and Yu, Minhui and Jiang, Bowen and Su, Zhan and others},
  journal={arXiv preprint arXiv:2603.11558},
  year={2026}
}

@article{xiao2026enpire,
  title={ENPIRE: Agentic Robot Policy Self-Improvement in the Real World},
  author={Xiao, Wenli and Xie, Jia and Zhang, Tonghe and Lin, Haotian and Fu, Letian and Xue, Haoru and Lu, Jalen and Yang, Yi and Dai, Cunxi and Wang, Zi and others},
  journal={arXiv preprint arXiv:2606.19980},
  year={2026}
}

@article{zhao2023learning,
  title={Learning fine-grained bimanual manipulation with low-cost hardware},
  author={Zhao, Tony Z and Kumar, Vikash and Levine, Sergey and Finn, Chelsea},
  journal={arXiv preprint arXiv:2304.13705},
  year={2023}
}

@article{chi2025diffusion,
  title={Diffusion policy: Visuomotor policy learning via action diffusion},
  author={Chi, Cheng and Xu, Zhenjia and Feng, Siyuan and Cousineau, Eric and Du, Yilun and Burchfiel, Benjamin and Tedrake, Russ and Song, Shuran},
  journal={The International Journal of Robotics Research},
  volume={44},
  number={10-11},
  pages={1684--1704},
  year={2025},
  publisher={Sage Publications Sage UK: London, England}
}

@inproceedings{carion2026sam,
  title={Sam 3: Segment anything with concepts},
  author={Carion, Nicolas and Gustafson, Laura and Hu, Yuan-Ting and Debnath, Shoubhik and Hu, Ronghang and Suris Coll-Vinent, Didac and Ryali, Chaitanya and Alwala, Kalyan Vasudev and Khedr, Haitham and Huang, Andrew and others},
  booktitle={International Conference on Learning Representations},
  volume={2026},
  pages={138846--138923},
  year={2026}
}

@article{intelligence2025pi_,
  title={$\pi_{0.5}$: a Vision-Language-Action Model with Open-World Generalization},
  author={Intelligence, Physical and Black, Kevin and Brown, Noah and Darpinian, James and Dhabalia, Karan and Driess, Danny and Esmail, Adnan and Equi, Michael and Finn, Chelsea and Fusai, Niccolo and others},
  journal={arXiv preprint arXiv:2504.16054},
  year={2025}
}

@article{kim2025fine,
  title={Fine-tuning vision-language-action models: Optimizing speed and success},
  author={Kim, Moo Jin and Finn, Chelsea and Liang, Percy},
  journal={arXiv preprint arXiv:2502.19645},
  year={2025}
}

@article{hu2026rac,
  title={Rac: Robot learning for long-horizon tasks by scaling recovery and correction},
  author={Hu, Zheyuan and Wu, Robyn and Enock, Naveen and Li, Jasmine and Kadakia, Riya and Erickson, Zackory and Kumar, Aviral},
  journal={IEEE Transactions on Robotics},
  year={2026},
  publisher={IEEE}
}

@article{luo2025precise,
  title={Precise and dexterous robotic manipulation via human-in-the-loop reinforcement learning},
  author={Luo, Jianlan and Xu, Charles and Wu, Jeffrey and Levine, Sergey},
  journal={Science Robotics},
  volume={10},
  number={105},
  pages={eads5033},
  year={2025},
  publisher={American Association for the Advancement of Science}
}

@inproceedings{han2026graspgen,
  title={GraspGen-X: Cross-Embodiment 6-DOF Diffusion-based Grasping},
  author={Han, Beining and Chao, Yu-Wei and Coumans, Erwin and Eppner, Clemens and Deng, Jia and Birchfield, Stan and Murali, Adithyavairavan},
  booktitle={Proceedings of the IEEE/CVF Conference on Computer Vision and Pattern Recognition},
  pages={20878--20889},
  year={2026}
}

@article{li2025nesypack,
  title={Nesypack: A neuro-symbolic framework for bimanual logistics packing},
  author={Li, Bowei and Yu, Peiqi and Tang, Zhenran and Zhou, Han and Sun, Yifan and Liu, Ruixuan and Liu, Changliu},
  journal={arXiv preprint arXiv:2506.06567},
  year={2025}
}

@inproceedings{ross2011reduction,
  title={A reduction of imitation learning and structured prediction to no-regret online learning},
  author={Ross, St{\'e}phane and Gordon, Geoffrey and Bagnell, Drew},
  booktitle={Proceedings of the fourteenth international conference on artificial intelligence and statistics},
  pages={627--635},
  year={2011},
  organization={JMLR Workshop and Conference Proceedings}
}

@article{yu2026maniskillformer,
  title={ManiSkillFormer: Demonstration-Free Compositional Manipulation via Task-Conditioned Geometric Contracts},
  author={Yu, Peiqi and Dabhi, Mosam and Li, Shangtao and Li, Bowei and Jeni, Laszlo and Liu, Changliu},
  journal={arXiv preprint arXiv:2609.16331},
  year={2026}
}

\clearpage

\useRomanappendicesfalse
\appendices
\setcounter{figure}{0}
\setcounter{table}{0}
\renewcommand{\thefigure}{S\arabic{figure}}
\renewcommand{\thetable}{S\arabic{table}}
\renewcommand{\theHfigure}{supp.\arabic{figure}}
\renewcommand{\theHtable}{supp.\arabic{table}}
\makeatletter
\setlength{\@fptop}{0pt}
\setlength{\@dblfptop}{0pt}
\newcommand{\suppfigurecaption}{\def\@captype{figure}\caption}
\makeatother

\newlength{\suppcolumnwidth}\setlength{\suppcolumnwidth}{\columnwidth}
\section{Policy Training}
\label{app:training}
Table~\ref{tab:supp-training} summarizes the training configurations
for the main real-robot experiments. Each task uses 1,000 successful
simulation demonstrations shared across architectures. The final
checkpoints are deployed without early stopping or validation-based selection.

Before training, the learning agent converts the shared demonstrations
into architecture-specific observation and action formats. The head and
two wrist cameras provide visual observations, and demonstration recording
and policy execution both operate at 10\,Hz. This shared source dataset
allows comparison across architectures without recollecting task demonstrations.

\begin{table}[H]
\caption{Training configurations for real-robot transfer.}
\label{tab:supp-training}
\centering
\footnotesize
\setlength{\tabcolsep}{3pt}
\renewcommand{\arraystretch}{1.10}
\begin{tabular}{@{}lccc@{}}
\toprule
 & ACT & DP & $\pi_{0.5}$ \\
\midrule
Initialization & ImageNet$^*$ & Scratch & \texttt{pi05\_base} \\
Training steps & 200,000 & 200,000 & 30,000 \\
Batch size & 128 & 128 & 32 \\
Optimizer & AdamW & Adam & AdamW \\
Peak learning rate & $10^{-5}$ & $10^{-4}$ & $2.5\!\times\!10^{-5}$ \\
LR schedule & Constant & Cosine & Cosine \\
Warm-up steps & --- & 500 & 1,000 \\
\bottomrule
\end{tabular}
\smallskip
\parbox{\columnwidth}{\footnotesize $^*$ACT uses an ImageNet-pretrained
ResNet-18 backbone; other parameters are initialized from scratch.
$\pi_{0.5}$ uses LoRA with non-LoRA parameters frozen.}
\end{table}

\section{Recovery Demonstration Generation}
\label{app:expert-recovery}
The training-free expert uses ground-truth simulation states to execute
state-based skills. After an arm displacement or induced failed grasp,
it replans from the disturbed state. Only successful expert continuations
are retained for training; the injected perturbations are excluded.
Figure~\ref{fig:supp-expert-recovery} illustrates both perturbations.

Perturbations alter the current execution state while preserving the task
objective. Recovery targets are recomputed from the updated object and
robot states, pairing disturbed observations with corrective expert actions.

\noindent\begin{minipage}{\columnwidth}
\centering
\includegraphics[width=\columnwidth]{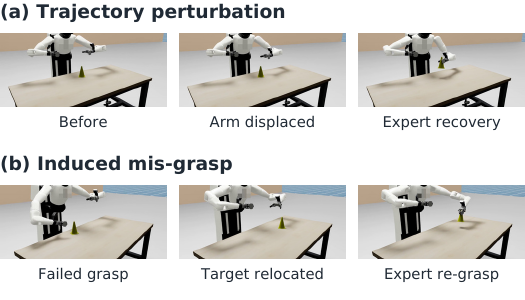}
\suppfigurecaption{Recovery data collection in simulation. The state-based expert
resumes the original task after (a) arm displacement or (b) an induced
mis-grasp and target relocation. These sequences precede policy training.}
\label{fig:supp-expert-recovery}
\end{minipage}\par\medskip

\newpage
\section{Data Scale}
\label{app:data-scale}
We train Diffusion Policy on nested subsets of 50, 100, 200, 500, and
1,000 screw-grasping demonstrations for 100,000 steps each. Evaluation
uses 40 held-out simulation episodes per policy.
Figure~\ref{fig:supp-data-scale} plots the main paper's results:
success increases from 0\% to 62.5\% over the evaluated dataset sizes.

The training-step budget is fixed across dataset sizes. Thus, the curve
measures performance with additional generated data under the same number
of optimization steps.

\noindent\begin{minipage}{\columnwidth}
\centering
\includegraphics[width=\columnwidth]{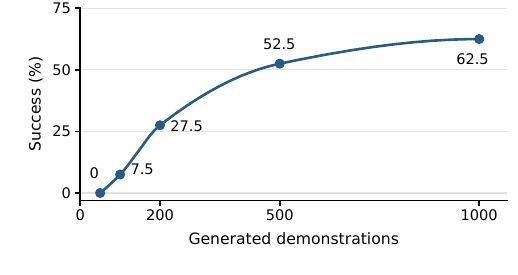}
\suppfigurecaption{Data scale with a fixed training budget. Values are
replotted from the main paper; the line is a visual guide.}
\label{fig:supp-data-scale}
\end{minipage}\par\medskip

\section{Real-Robot Evaluation and Recovery}
\label{app:real-recovery}
We also test recovery when an operator displaces the target during real
execution. The learned visuomotor policy uses updated observations to
re-grasp the object and complete the task.
Figure~\ref{fig:supp-real-recovery} provides additional examples beyond
the cone sequence in the main paper.

These selected recordings illustrate recovery behavior; they are separate
from the 40-trial task--policy success-rate evaluation.

\noindent\begin{minipage}{\columnwidth}
\centering
\includegraphics[width=\columnwidth]{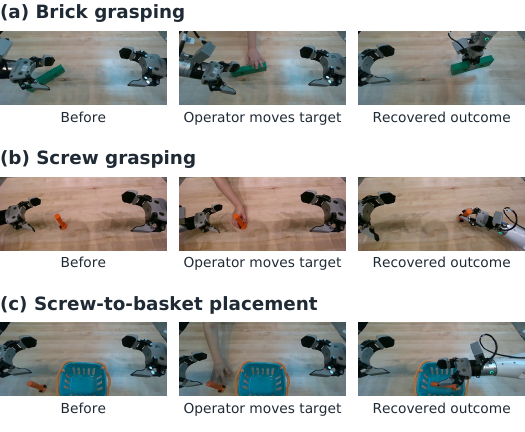}
\suppfigurecaption{Real-world recovery from target displacement. Each row preserves
the order of the initial state, operator intervention, and recovered
outcome. These selected sequences are qualitative examples.}
\label{fig:supp-real-recovery}
\end{minipage}\par\medskip

\clearpage
\twocolumn[{
\begin{minipage}{\textwidth}
\centering
\includegraphics[width=\textwidth]{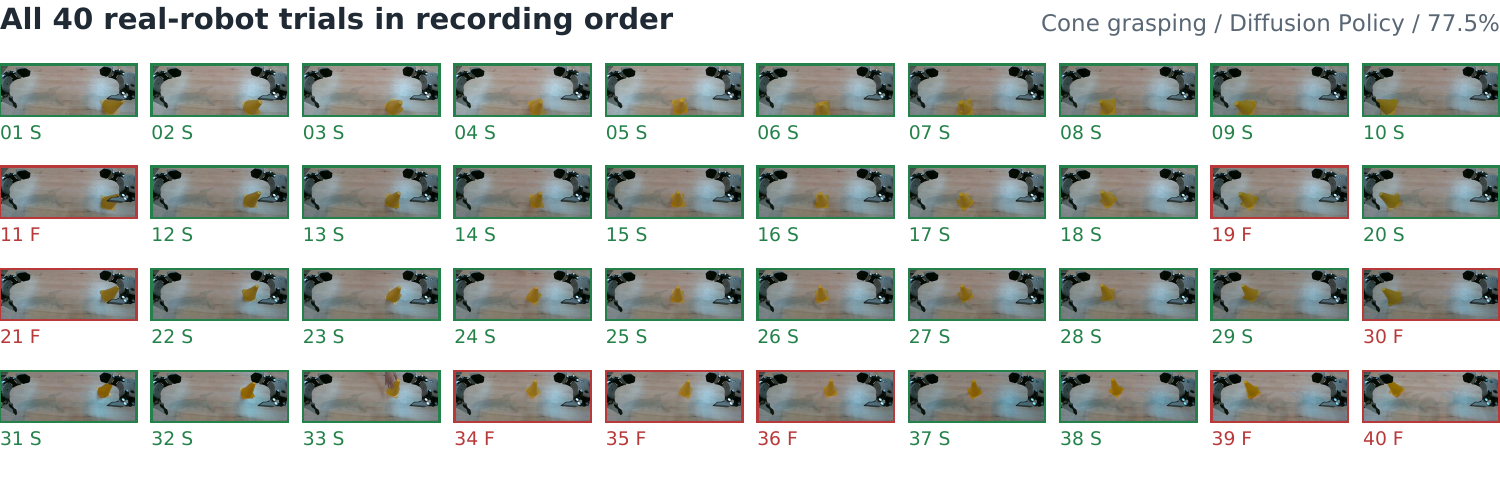}
\suppfigurecaption{Initial frames and outcomes for all 40 cone-grasping trials.
S/green denotes success and F/red denotes failure. Trial ordering matches
the corresponding recording list; failures are retained.}
\label{fig:supp-trial-grid}
\end{minipage}
\vspace{0.6em}
}]

\renewcommand{\thesubsection}{\thesection.\arabic{subsection}}
\renewcommand{\thesubsectiondis}{\thesection.\arabic{subsection}}
\subsection{Evaluation Sampling}
\label{app:evaluation-sampling}
Before each real-robot trial, object placements are resampled within the
task workspace. Figure~\ref{fig:supp-placement-map} overlays initial cone
placements; Fig.~\ref{fig:supp-trial-grid} includes all 40 trials and their
outcomes. This Diffusion Policy trial set contains 31 successes and nine
failures (77.5\%). Thumbnail positions index recordings rather than
physical sampling locations.

For grasping, targets are sampled uniformly over the workspace. For
regional sampling, each object is placed within its assigned region,
following the collection protocol.

\noindent\begin{minipage}{\columnwidth}
\centering
\includegraphics[width=0.95\columnwidth]{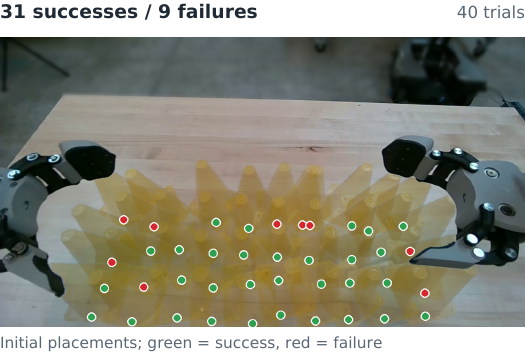}
\suppfigurecaption{Sampled initial placements for cone grasping. Overlaid silhouettes
represent different trials, not the trajectory of one moving object.}
\label{fig:supp-placement-map}
\end{minipage}\par\medskip

\section{Tool-Feedback-Driven Workflow Recovery}
\label{app:agent-recovery}
With the same input and tools, the fixed pipeline retries the flat-nut
scene and aborts. ARSTAG uses grasp diagnostics to jointly revise the
simulated nut's scale, orientation, mass, and support contacts before
resuming collection
(Figs.~\ref{fig:supp-agent-recovery}--\ref{fig:supp-agent-visuals}).
The nut is enlarged by $1.5\times$ and rotated $90^\circ$ into an
upright pose; its mass is set to 0.02\,kg and support penetration is
corrected. The revised scene yields three expert demonstrations; a
subsequent 300-step policy records 0/2 simulation evaluation successes.

\newpage
\noindent\begin{minipage}{\columnwidth}
\centering
\includegraphics[width=0.88\columnwidth]{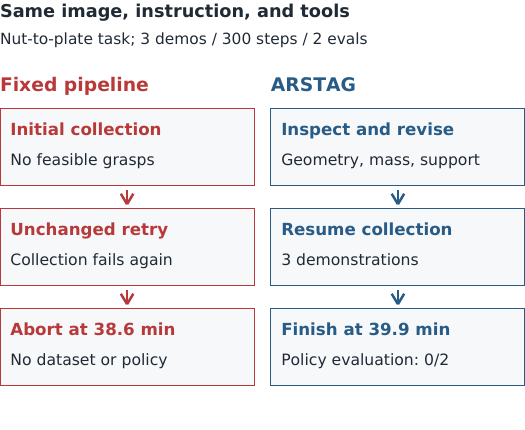}
\suppfigurecaption{Recorded workflow recovery. Initial-scene revision restores
expert data collection. The short training run tests workflow execution,
not converged-policy performance.}
\label{fig:supp-agent-recovery}
\end{minipage}\par\medskip

\noindent\begin{minipage}{\columnwidth}
\centering
\includegraphics[width=0.88\columnwidth]{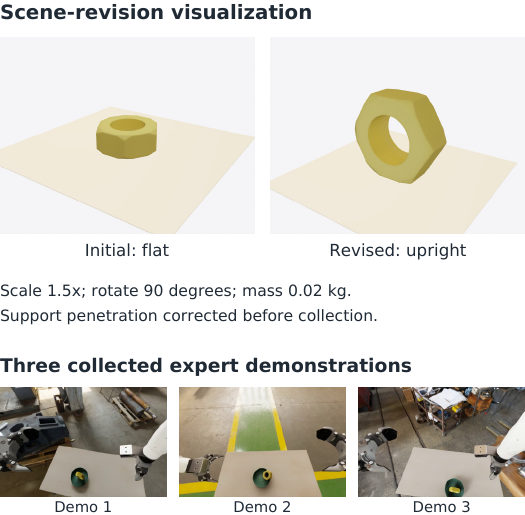}
\suppfigurecaption{Nut-to-plate scene revision and expert demonstrations. Upper:
illustrative renderings of the initial flat pose and revised upright
pose. Lower: final frames of three expert demonstrations collected
in the revised scene.}
\label{fig:supp-agent-visuals}
\end{minipage}\par\medskip

\clearpage
\twocolumn[{
\begin{minipage}{\textwidth}
\noindent\begin{minipage}[t]{\suppcolumnwidth}
\section{Support Repair}
\label{app:support}
Figure~\ref{fig:supp-support} compares settling with and without final
vertical support alignment, retaining root leveling, ground alignment,
and correction propagation in both conditions. The three selected examples extend
the main paper's quantitative evaluation to individual objects and
multi-object arrangements. Residual motion can remain on concave supports.

\end{minipage}\hfill\begin{minipage}[t]{\suppcolumnwidth}
\section{Instance-Level Identity Binding}
\label{app:binding}
Scene-graph captions prompt segmentation, and identity binding uses
appearance and spatial relations to match graph objects to candidate
masks. Figure~\ref{fig:supp-binding} shows two complementary cases involving
repeated classes, background distractors, and supporting surfaces.
Partial one-to-one matching prevents distinct objects from sharing a mask.

\end{minipage}\par\vspace{0.7em}
\centering
\includegraphics[width=\textwidth]{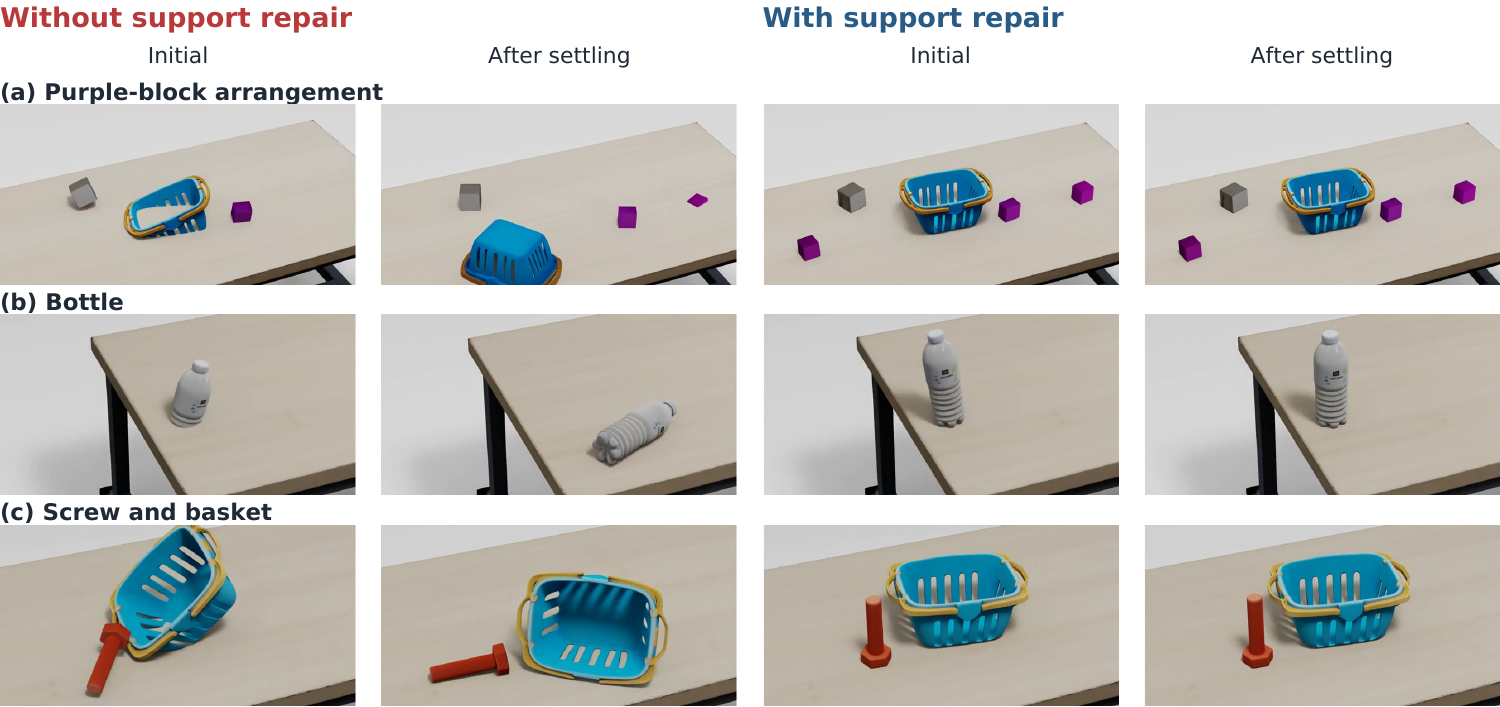}
\suppfigurecaption{Selected support-repair comparisons with pronounced settling motion
without repair. Each row shows initial and later states over the same
video interval in both conditions; intervals differ across scenes.}
\label{fig:supp-support}
\par\vspace{1.2em}
\centering
\includegraphics[width=\textwidth]{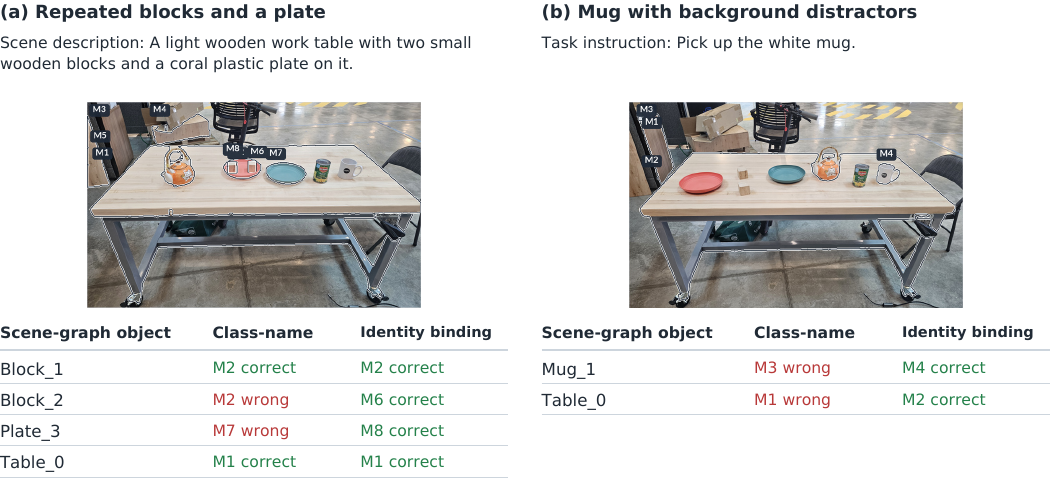}
\suppfigurecaption{Identity binding resolves repeated-object and background-distractor ambiguity. Assignments are copied from the recorded annotations; mask identifiers are local to each scene. Panel (a) uses a scene description, and (b) uses a task instruction.}
\label{fig:supp-binding}
\end{minipage}
\vspace{0.6em}
}]

\end{document}